\documentclass[runningheads]{llncs}

\usepackage{eccv}

\usepackage{eccvabbrv}

\usepackage{graphicx}
\usepackage{booktabs}

\usepackage[accsupp]{axessibility}  

\usepackage{hyperref}

\usepackage{orcidlink}

\usepackage{enumitem}
\usepackage{makecell}
\usepackage{multirow}
\usepackage{pifont}
\usepackage{bbm}
\usepackage[table,xcdraw]{xcolor}
\usepackage[misc]{ifsym}
\usepackage{tcolorbox}
\tcbuselibrary{skins, breakable}

\begin{document}

\title{Linguistically-Aligned and Visually-Grounded Preference Optimization for Clinically-Augmented Medical Report Generation}
\def\thefootnote{$\dag$}\footnotetext{Equal~contribution; \textrm{\Letter}~Corresponding author.\\
}
\titlerunning{Linguistically-Aligned and Visually-Grounded Preference Optimization}

\author{Qiang Hu$^{\dag}$\orcidlink{0000-0001-8209-3754} \and
Yuxuan Luo$^{\dag}$\orcidlink{0009-0008-1719-9454} \and
Yingjie Guo\orcidlink{0009-0008-9857-9001} \and
Hao Wang\orcidlink{0009-0005-6644-6253} \and
Qimei Wang\orcidlink{0009-0005-5697-3252} \and\\
Qiang Li\orcidlink{0000-0002-9815-4432} \and
Zhiwei Wang\textsuperscript{(\Letter)}\orcidlink{0000-0002-1612-8573}
}

\authorrunning{Q. Hu et al.}

\institute{Huazhong University of Science and Technology\\
\email{\{huqiang77,zwwang\}@hust.edu.cn}}

\maketitle
\begin{abstract}
Despite significant advances in Medical Report Generation (MRG), the reliability remains constrained by the prevalence of factual errors.
While Direct Preference Optimization (DPO) has emerged as a promising post-training paradigm to enhance the performance of Supervised Fine-Tuned (SFT) MRG models, existing DPO-based MRG methods typically adopt a naive preference construction that directly pairs model-generated reports with ground truth reports.
This strategy inadvertently entangles critical clinical findings with clinically irrelevant linguistic characteristics, and fundamentally lacks explicit vision-language alignment.
To address these challenges, we propose DPO-Clin, a novel post-training framework that focuses preference optimization on clinical findings and cross-modal alignment.
First, we introduce the Entity-level Clinical Diagnostic (ECD) module to perform a precise entity-level factual diagnosis. ECD guides the generation of linguistically-aligned report preference pairs, isolating clinical discrepancies from linguistic variations.
Second, to achieve fine-grained cross-modal alignment, we develop M$^2$DPO, a retrieval-augmented multi-modal DPO variant that enforces textual preference inversion triggered by visual context switches.
Third, we locate correct yet highly uncertain predicted entities and apply counterfactual modifications to construct targeted preference data for latent risk mitigation, thereby further enhancing the model reliability.
Extensive experiments on two public chest X-ray datasets (MIMIC-CXR and IU X-Ray) and an in-house endoscopy dataset demonstrate that DPO-Clin significantly improves the SFT baselines on clinical-aware metrics. Furthermore, it achieves superior performance over existing DPO-based MRG methods, exhibiting robust generalizability across distinct baseline architectures and diverse medical imaging modalities.
Codes are available at \url{https://github.com/Huster-Hq/DPO-Clin}.
\end{abstract}
\section{Introduction}
Automatic Medical Report Generation (MRG) aims to translate medical images into comprehensive textual reports that summarize clinically meaningful findings. By enabling scalable and standardized interpretation of medical imaging data, MRG holds significant potential to improve diagnostic efficiency, reduce clinician workload, and enhance consistency in clinical decision-making.

Early researches in MRG primarily focused on architectural innovations tailored to vision-language modeling. Representative efforts explored multi-modal attention mechanisms~\cite{liu2021contrastive,chen2020generating}, region-aware feature localization~\cite{tanida2023interactive,chen2025large}, and the incorporation of structured medical knowledge~\cite{zhang2020radiology,li2023dynamic}.
Instead of designing specialized architectures from scratch, current state-of-the-art approaches increasingly adapt general-purpose Large Language Models (LLMs) to medical report generation~\cite{jin2024promptmrg,wang2023r2gengpt,hou2025radar}. However, since they are trained on broad-domain data, effective deployment in the medical domain requires domain-specific adaptation.

The most straightforward adaptation strategy is Supervised Fine-Tuning (SFT), where models are optimized to reproduce reference reports using token-level supervision (\textit{e.g.}, cross-entropy loss).
Despite its simplicity and technical maturity, SFT fundamentally optimizes imitation rather than clinical preference understanding. It fails to capture why one report is clinically superior to another and struggles when multiple valid descriptions coexist but differ in diagnostic utility, a challenge commonly characterized as the lack of preference alignment.

To address this limitation, recent studies adopt post-training strategies following SFT. A dominant paradigm is Reinforcement Learning from Human Feedback (RLHF)~\cite{hein2024preference,zhou2024large,wang2025beyond}, which introduces an auxiliary reward model to evaluate generated reports. The generation model is then optimized to maximize expected rewards through reinforcement learning algorithms such as PPO~\cite{schulman2017proximal} or GRPO~\cite{shao2024deepseekmath}. Although effective, RLHF suffers from substantial computational overhead, training instability, and potential reward mis-specification, which may lead to reward hacking and clinically unreliable optimization.

Recently, Direct Preference Optimization (DPO)~\cite{rafailov2023direct} has emerged as a simpler and more stable alternative to RLHF by eliminating explicit reward modeling and policy-gradient optimization. Given pairwise preference data consisting of a prompt, a preferred response, and a dispreferred response, DPO directly encourages the model to assign higher likelihood to the preferred response, enabling efficient one-step preference alignment.

\begin{figure}[t!]
    \centering
    \includegraphics[width=\linewidth]{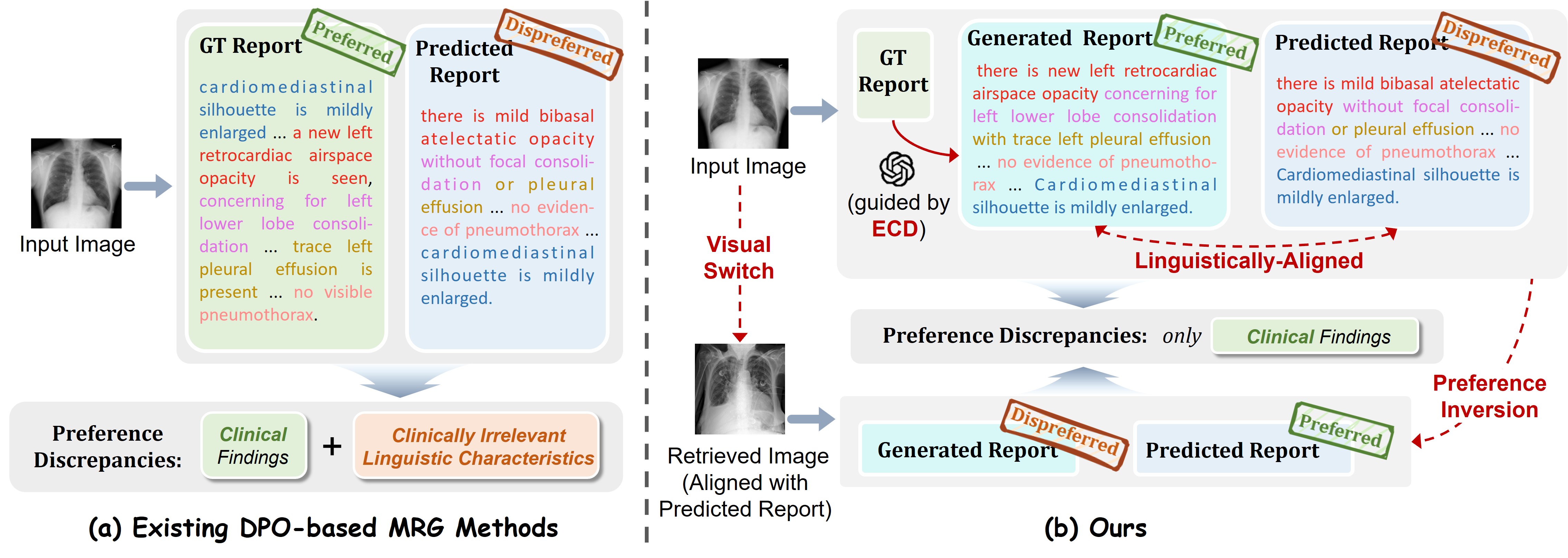}
    \caption{Comparison between existing DPO-based MRG methods and our DPO-Clin. We utilize distinct colors to indicate descriptions corresponding to different clinical findings. (a) Existing methods construct report preference pairs directly from raw GT and predicted reports, inadvertently entangling critical clinical findings with clinically irrelevant linguistic variations. (b) Guided by the proposed Entity-level Clinical Diagnostic (ECD) module, our DPO-Clin isolates clinical discrepancies by generating linguistically-aligned preference reports. Furthermore, DPO-Clin enforces fine-grained vision-language alignment through a visual-context-triggered preference inversion mechanism. For visual clarity, the mechanism for mitigating latent risks is omitted.}
    \label{fig:introduction}
\end{figure}

Existing DPO-based MRG approaches~\cite{zhummedpo} typically construct preference pairs by treating ground-truth (GT) reports as preferred responses and model-generated outputs as dispreferred ones, followed by uniform optimization over all tokens. However, as shown in Fig.~\ref{fig:introduction}(a), this paradigm overlooks the fundamental nature of medical reports: critical findings, such as disease entities and their assertion status, are tightly interwoven with clinically irrelevant linguistic characteristics, including connective phrases, syntactic variations, and word ordering. Assigning equal importance to all tokens dilutes the alignment signal and weakens supervision on diagnostically relevant information, encouraging the model to favor linguistic similarity over clinical correctness.

Moreover, existing methods expose the model only to the positive image corresponding to the GT report during preference alignment, without introducing contrastive visual references for the same report. In clinical practice, diagnostic conclusions often depend on subtle and localized visual differences across images~\cite{yang2025spatio,hu2026pairing}. Without contrasting image-level preferences, models struggle to associate textual descriptions with discriminative visual evidence, which may ultimately lead to clinically significant errors such as false positives, false negatives.

To address these challenges, we propose \textbf{DPO-Clin}, a \textbf{DPO}-based framework that explicitly isolates \textbf{clin}ically relevant content and establishes multi-modal-aware optimization.
As illustrated in Fig.~\ref{fig:introduction}(b), our core insight lies in constructing preference report pairs that contain exclusively clinical discrepancies.
To this end, we introduce an \textbf{E}ntity-level \textbf{C}linical \textbf{D}iagnostic (\textbf{ECD}) module to perform precise entity-level diagnostic.
The diagnostic results guide LLM (GPT-4o~\cite{achiam2023gpt}) to generate a preferred report that is linguistically-aligned with the predicted report, while strictly adhering to the clinical facts of the GT report. Pairing this generated report with the original prediction forces the model optimization to focus on clinically relevant discrepancies.
Furthermore, we propose $\mathbf{M^2DPO}$, a retrieval-augmented \textbf{M}ulti-\textbf{M}odal \textbf{DPO} variant that enforces a textual preference inversion triggered by visual context switches. This design tightly couples visual discrepancies with textual factual differences, compelling the model to ground diagnostic assertions in discriminative pathological cues rather than spurious correlations.
Beyond mitigating these explicit textual errors, we also emphasize eliminating latent risks—medical entities that are predicted correctly yet with high uncertainty. We localize these uncertain entities and apply counterfactual modifications to construct additional preference data, thereby further enhancing the overall clinical reliability of the MRG model.

Our main contributions are summarized as follows:
\begin{itemize}[itemsep=2pt, parsep=0pt, topsep=0pt, wide=5pt]
    \item We propose DPO-Clin, an innovative DPO-based post-training framework tailored for MRG. By explicitly directing preference optimization to focus on clinically relevant multi-modal discrepancies, DPO-Clin effectively mitigates both explicit textual errors and latent clinical risks inherent in SFT baselines, significantly improving report generation quality.
    \item We introduce the ECD module for precise entity-level diagnosis, guiding the construction of reliable, linguistically-aligned report preference pairs. Furthermore, we develop M$^2$DPO, which enforces textual preference inversion triggered by visual context switches to achieve fine-grained vision-language alignment. Additionally, we systematically localize uncertain entities and apply counterfactual modifications to construct targeted preference data for latent risk mitigation.
    \item We conduct comprehensive evaluations across two chest X-ray datasets (MIMIC-CXR and IU X-Ray) and an in-house endoscopy dataset. Extensive results demonstrate that DPO-Clin significantly elevates the clinical efficacy of SFT baselines and achieves superior performance compared to other DPO-based MRG methods, exhibiting remarkable generalizability across diverse baseline architectures (R2GenGPT and RADAR) and distinct medical imaging modalities.
\end{itemize}
\section{Related Works}
\subsection{Medical Report Generation}
Medical Report Generation (MRG) aims to automatically generate coherent and accurate diagnostic reports from medical images.
Early works primarily focused on enhancing report quality through architectural innovations.
These include designing advanced attention mechanisms to capture complex cross-modal interactions~\cite{chen2020generating,wang2023metransformer}, introducing region-aware localization priors to explicitly ground text in specific pathological regions~\cite{chen2025large,tanida2023interactive,hu2024sali}, and integrating structured medical knowledge~\cite{zhang2020radiology,yang2022knowledge,li2023dynamic} or auxiliary classification networks~\cite{jing2018automatic,fan2024medical,hu2025holistic,bu2024instance} to guide decoding and rectify factual deviations.
Recently, an advanced paradigm has gained increasing attention, which adapts general-purpose Large Language Models (LLMs) to MRG via Supervised Fine-Tuning (SFT)~\cite{hou2025radar,jin2024promptmrg,wang2023r2gengpt}.
Benefiting from massive pre-training corpora, this paradigm preserves fluent text generation capabilities and achieves leading performance.
Nevertheless, SFT inherently applies uniform token-level supervision, preventing the model from focusing exclusively on clinically significant content. As a result, the generation process remains plagued by prevalent factual errors, undermining the reliability.

\subsection{Direct Preference Optimization}
Direct Preference Optimization (DPO) has emerged as an efficient post-training technique, widely adopted to enhance the capabilities of the SFT baseline.
Based on the preference data curation strategy, these approaches can generally be categorized into two types: those relying on manually annotated preference pairs~\cite{yu2024rlhf,yang2024using,hong2024adaptive}, and those utilizing heuristic rules or automatic evaluation models to generate large-scale preference data~\cite{hong2024adaptive,zhang2024automated,yu2025rlaif,wang2025enhancing}.
Inspired by the success in general domains, recent studies~\cite{sun2024self,banerjee2024direct,liang2025chexpo} have explored the adaptability of DPO in medical tasks, including MRG.
For example, RRG-DPO~\cite{liu2025rrg} automatically retrieves paired preference reports from the training database. MMedPO~\cite{zhummedpo} adjusts optimization weights by quantifying the clinical relevance scores of preference reports, and \cite{hein2024preference} employs report-exclusive evaluation metrics to guide the preference prioritization among multiple model predictions.
However, these methods overlook the intrinsic characteristic of medical reports, where critical clinical findings are heavily intertwined with clinically irrelevant linguistic variations.
Directly applying DPO to preference reports fails to capture clinically relevant textual discrepancies.
Furthermore, they lack sufficient exploration in leveraging DPO to drive fine-grained vision-language alignment, primarily executing a text-centric optimization.
\section{Preliminary on DPO-based MRG}
Direct Preference Optimization (DPO)~\cite{rafailov2023direct} has emerged as a simple yet effective paradigm for aligning generative models with human preferences in a stable and reward-free manner. It directly optimizes the policy to favor preferred completions over dispreferred ones relative to a frozen reference policy. Given a query medical image $x$, current DPO-based MRG approaches~\cite{liu2025rrg} typically treat incorrect model predictions $y^{\text{pre}}$ as dispreferred reports and ground-truth (GT) reports $y^{\text{GT}}$ as preferred ones, maximizing the log-odds margin between them under the target policy $\pi_\theta$ after subtracting the corresponding log-odds under a reference policy $\pi_{\text{ref}}$:
\begin{equation}
\small
\mathcal{L}_{\text{DPO}}(x,y^{\text{GT}},y^{\text{pre}})=-\log\sigma\left(\beta \left[\log\frac{\pi_{\theta}(y^{\text{GT}}|x)}{\pi_{\text{ref}}(y^{\text{GT}}|x)}-\log\frac{\pi_{\theta}(y^{\text{pre}}|x)}{\pi_{\text{ref}}(y^{\text{pre}}|x) }\right]\right),
\label{eq:standard_DPO}
\end{equation}
where $\beta$ is a temperature hyperparameter controlling the strength of preference alignment and $\sigma(\cdot)$ denotes the sigmoid function.

However, this paradigm suffers from three limitations that are particularly critical in MRG.
\textbf{First}, the preference discrepancies between $y^{\text{pre}}$ and $y^{\text{GT}}$ entangle critical clinical findings with clinically irrelevant linguistic variations.
Such uniform, report-level preference assignments prevent the optimization process from isolating clinically relevant textual discrepancies.
\textbf{Second}, enforcing textual preference supervision conditioned solely on a static image fails to compel the model to ground textual discrepancies in specific pathological visual regions, thereby hindering fine-grained vision-language alignment.
\textbf{Third}, it exclusively focuses on explicit textual factual errors, ignoring the optimization of `latent risks'—correct yet highly uncertain predictions.

\section{Methodology}
\subsection{Overview}
\begin{figure}[t]
    \centering
    \includegraphics[width=\linewidth]{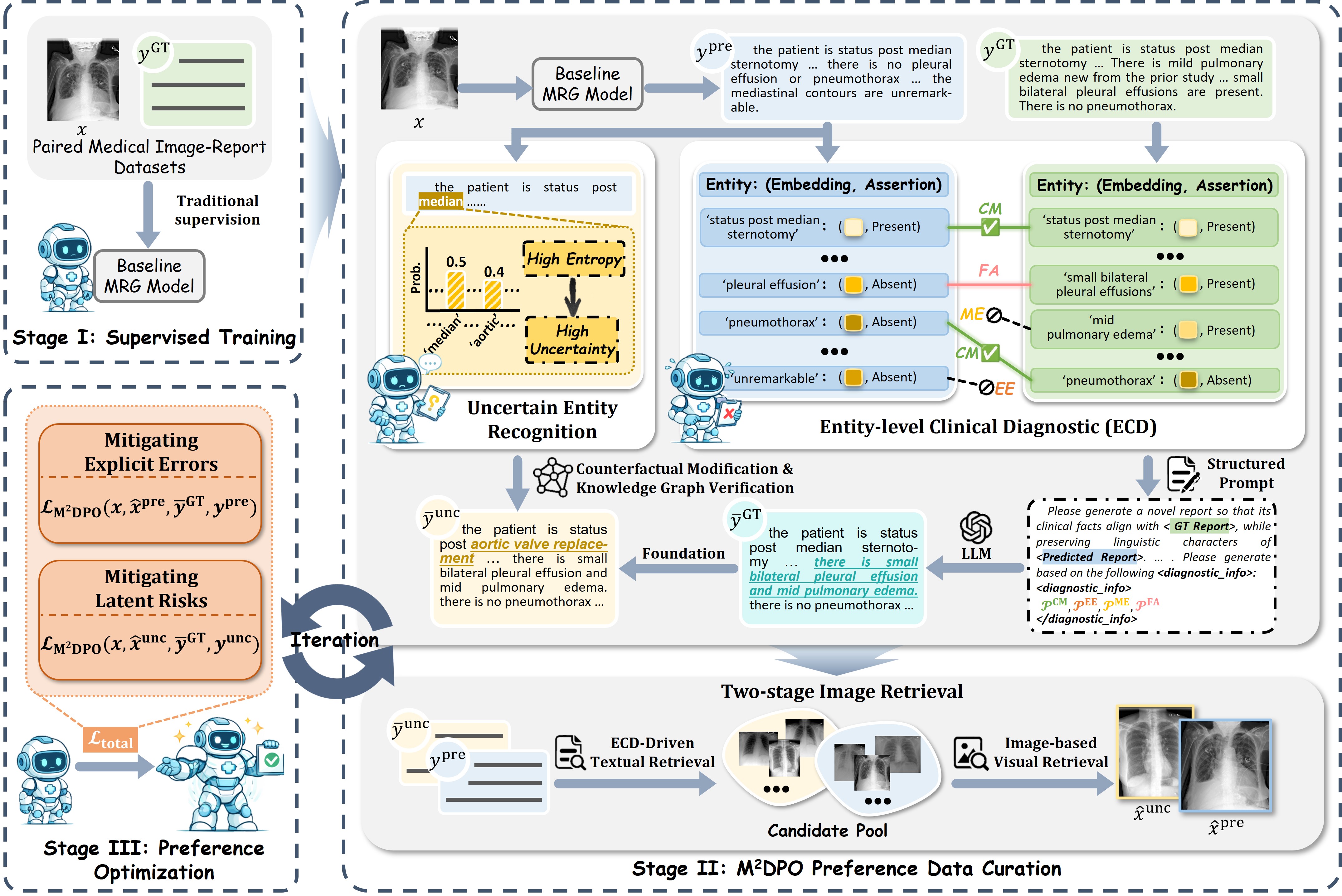}
    \caption{Overview of the proposed DPO-Clin. DPO-Clin is a DPO-based post-training framework designed to optimize the SFT MRG baseline. It leverages an ECD module to assist LLM in generating linguistically-aligned preferred reports $\overline{y}^{\text{GT}}$, and applies uncertain entity recognition to obtain counterfactual reports $\overline{y}^{\text{unc}}$. By incorporating a two-stage image retrieval strategy, it constructs multi-modal preference data for M$^2$DPO, which ultimately achieves robust clinical preference alignment.}
    \label{fig:architecture}
\end{figure}
As illustrated in Fig.~\ref{fig:architecture}, we propose {DPO-Clin}, a post-training framework tailored for direct clinical preference alignment.
DPO-Clin introduces three core components to tackle the aforementioned limitations.
First, we leverage an Entity-level Clinical Diagnostic (ECD) module to assist the Large Language Model (LLM) in precisely eliminating clinically irrelevant linguistic variations from the preference report pairs.
Second, we propose M$^2$DPO, a novel multi-modal variant of DPO, to enforce a textual preference inversion triggered by visual context switches, thereby achieving fine-grained vision-language alignment.
Third, we localize correct yet highly uncertain predicted clinical entities and apply counterfactual modifications to construct the dispreferred report, optimizing the model's reliability against latent risks.
The following sections elaborate on each component and the overall training pipeline.

\subsection{ECD-Assisted Linguistically-Aligned Preference Optimization}
\label{sec:ECD_Assisted_Optimization}
Medical reports intrinsically intertwine two types of content: \textit{clinical findings}, \textit{i.e.}, medical entities and their assertion status (Present/Absent), and \textit{clinically irrelevant linguistic characteristics}, \textit{i.e.}, connective phrases, syntactic variations, and word ordering.
To construct linguistically-aligned preference report pairs, an intuitive approach is to prompt an LLM to generate a new preferred report that preserves the linguistic characteristics of the prediction report, while adhering to the clinical facts of the GT report.
However, existing general-purpose LLMs typically exhibit limited sensitivity to medical entities and struggle with medical-specific semantic comprehension, rendering their direct generation unreliable.

To this end, we introduce Entity-level Clinical Diagnostic (ECD) module that performs fine-grained entity-level analysis.
Specifically, ECD employs the RaTE-NER model~\cite{zhao2024ratescore} to parse both the predicted report $y^{\text{pre}}$ and the GT report $y^{\text{GT}}$, extracting medical entities $\mathbf{e}$ along with their corresponding semantic embeddings $\mathbf{E}$ and assertion statuses $\mathbf{A} \in \{\text{Present},\text{Absent}\}$.
This yields two structured sets: $\mathcal{S}^{\text{pre}} = \{ \mathbf{e}^{\text{pre}}_i~\text{:}~( \mathbf{E}^{\text{pre}}_i, \mathbf{A}^{\text{pre}}i) \}_{i=1}^{N}$ and $\mathcal{S}^{\text{GT}} = \{ \mathbf{e}^{\text{GT}}_j~\text{:}~( \mathbf{E}^{\text{GT}}_j, \mathbf{A}^{\text{GT}}_j) \}_{j=1}^{M}$, where $N$ and $M$ denote entity counts in the predicted and GT reports, respectively.

Following this, to establish fine-grained correspondences, we formulate entity matching as a threshold-constrained linear assignment problem.
The matching cost between the $i$-th predicted entity and $j$-th GT entity is defined as their cosine distance: ${\text{cost}}_{i,j} = 1 - \cos(\mathbf{E}^{\text{pre}}_i, \mathbf{E}^{\text{GT}}_j)$. The optimal assignment matrix $\mathbf{M}^*$ minimizes the global matching cost:
\begin{equation}
\label{eq:matching}
\small
\mathbf{M}^* = \underset{\mathbf{M} \in \{0,1\}^{N \times M}}{\arg\min} \sum_{i=1}^{N} \sum_{j=1}^{M} \text{cost}_{i,j} \cdot \mathbf{M}_{i,j},~~
\text{s.t.}\sum_{j} \mathbf{M}_{i,j} = \{0, 1\},\sum_{i} \mathbf{M}_{i,j} = \{0,1\}.
\end{equation}
A similarity threshold $\tau$ further constrains the matching, requiring $\text{cost}_{i,j} < 1 - \tau$ for any matched pair $(\mathbf{e}_i^{\text{pre}},\mathbf{e}_j^{\text{GT}})$.
Based on $\mathbf{M}^*$, we categorize entities into four clinically meaningful subsets, each corresponding to a distinct error type that will guide subsequent preference construction:
\begin{itemize}[itemsep=2pt, parsep=0pt, topsep=0pt, wide=5pt]
\item \textbf{Correct Matches} ($\mathcal{P}^{\text{CM}}$): Matched entity pairs $(\mathbf{e}_i^{\text{pre}}, \mathbf{e}_j^{\text{GT}})$ with identical assertion statuses ($\mathbf{M}_{i,j}^{*}=1$ and $\mathbf{A}_i=\mathbf{A}_j$), denoting accurate predictions.
\item \textbf{Extraneous Entities} ($\mathcal{P}^{\text{EE}}$): Unmatched predicted entities $\mathbf{e}_i^{\text{pre}}$ ($\sum_{j} \mathbf{M}^{*}_{i,j}=0$), indicating fabricated findings that are absent in the GT report.
\item \textbf{Missing Entities} ($\mathcal{P}^{\text{ME}}$): Unmatched GT entities $\mathbf{e}_j^{\text{GT}}$ ($\sum_{i} \mathbf{M}^{*}_{i,j}=0$), representing clinically significant findings that the model ignored.
\item \textbf{False Assertions} ($\mathcal{P}^{\text{FA}}$): Matched entity pairs $(\mathbf{e}_i^{\text{pre}}, \mathbf{e}_j^{\text{GT}})$ with conflicting assertion statuses ($\mathbf{M}_{i,j}^{*}=1$ and $\mathbf{A}_i \neq \mathbf{A}_j$), capturing instances where the model correctly identifies an anatomical entity but misattributes its clinical state.
\end{itemize}

Finally, these fine-grained diagnostic details are compiled into a structured prompt, guiding the LLM (GPT-4o~\cite{achiam2023gpt}) to generate $\overline{y}^{\text{GT}}$, which strictly adheres to the clinical facts of $y^{\text{GT}}$ while preserving the linguistic characteristics of $y^{\text{pre}}$.
By pairing $\overline{y}^{\text{GT}}$ and $y^{\text{pre}}$ to construct the preference data, we ensure that the optimization process focuses exclusively on clinically relevant textual discrepancies.
This text-centric preference optimization objective is formulated as $\mathcal{L}_{\text{DPO}}(x, \overline{y}^{\text{GT}}, y^{\text{pre}})$.

\subsection{Retrieval-Augmented M$^2$DPO for Vision-Language Alignment}
\label{sec:method_image_retrival}
Recognizing the inherent multi-modal nature of MRG, it is crucial to optimize not only the textual output but also the model's visual grounding capabilities.
To this end, we propose $\mathbf{M^2DPO}$, a novel multi-modal variant of DPO.
Operating on a quadruplet $(x_1, x_2, y_1, y_2)$ where $y_1$ and $y_2$ are factually aligned with $x_1$ and $x_2$ respectively, M$^2$DPO enforces a textual preference inversion triggered by visual context switches.
Specifically, $y_1$ serves as the preferred report over $y_2$ under the visual context of $x_1$, whereas this preference order is inverted when the visual context switches to $x_2$.
Building upon the standard DPO objective defined in Eq.~(\ref{eq:standard_DPO}), the M$^2$DPO objective is formulated as follows:
\begin{equation}
\small
\mathcal{L}_{\text{M$^2$DPO}}(x_1,x_2,y_1,y_2)= \mathcal{L}_{\text{DPO}}(x_1,y_1,y_2) + \mathcal{L}_{\text{DPO}}(x_2, y_2, y_1).
\end{equation}
This objective enforces the model to ground clinical textual discrepancies in their corresponding local pathological visual discrepancies, thereby achieving fine-grained vision-language alignment.

To implement the M$^2$DPO objective, it is necessary to augment the preference triplets $(x, \overline{y}^{\text{GT}}, y^{\text{pre}})$ constructed in Sec.~\ref{sec:ECD_Assisted_Optimization} into quadruplets.
To this end, we employ a two-stage strategy to retrieve an image $\hat{x}^{\text{pre}}$ from the training database that is factually aligned with the predicted report $y^{\text{pre}}$. The detailed retrieval protocol is outlined below:
\begin{enumerate}[itemsep=2pt, parsep=0pt, topsep=0pt, wide=5pt, label=\textbf{\arabic*.}]
\item \textbf{ECD-Driven Textual Retrieval}: We leverage the proposed ECD module to evaluate the factual alignment between $y^{\text{pre}}$ and candidate reports in the training set. Specifically, we construct a candidate pool comprising images whose paired reports yield a valid set of Correct Matches ($|\mathcal{P}^{\text{CM}}| > 0$) while strictly maintaining empty error subsets ($\mathcal{P}^{\text{EE}} \cup \mathcal{P}^{\text{ME}} \cup \mathcal{P}^{\text{FA}} = \varnothing$) when diagnosed against $y^{\text{pre}}$. This entity-level filtering guarantees absolute clinical equivalence.
\item \textbf{Image-based Visual Retrieval}: Within the candidate pool, we extract spatial embeddings using MedKLIP~\cite{wu2023medklip}, and select the image with the highest cosine similarity to the query image $x$ as the final retrieved image $\hat{x}^{\text{pre}}$.
\end{enumerate}

This two-stage protocol efficiently ensures that $\hat{x}^{\text{pre}}$ is clinically congruent with $y^{\text{pre}}$ while minimizing clinically irrelevant visual shifts from $x$.
Thus, we construct the preference quadruplet $(x, \hat{x}^{\text{pre}}, \overline{y}^{\text{GT}}, y^{\text{pre}})$, and then augment the text-centric foundation $\mathcal{L}_{\text{DPO}}(x, \overline{y}^{\text{GT}}, y^{\text{pre}})$ in Sec.~\ref{sec:ECD_Assisted_Optimization} into the multi-modal objective $\mathcal{L}_{\text{M$^2$DPO}}(x, \hat{x}^{\text{pre}}, \overline{y}^{\text{GT}}, y^{\text{pre}})$.

\subsection{Counterfactual Data Construction for Latent Risk Mitigation}
\label{sec:method_counterfactual_data}
Beyond optimizing for explicit textual errors, it is also crucial to mitigate latent risks, defined as clinically relevant predictions that are factually correct yet highly uncertain.
To localize these risks, we propose an uncertain entity recognition strategy.
It quantifies the uncertainty of an entity by identifying the maximum token-level entropy among its constituent tokens.
Assuming an entity $\mathbf{e}$ consists of $K$ tokens, its uncertainty score is defined as:
\begin{equation}
\small
\mathcal{H}(\mathbf{e}|x) = \max_{k=1}^{K} \big( - \sum_{w \in \mathcal{V}} \pi_{\text{ref}}(w | {x}, \mathcal{T}_{<k}) \log \pi_{\text{ref}}(w | {x}, \mathcal{T}_{<k}) \big),
\label{eq:uncertainty_score}
\end{equation}
where $w$ denotes a candidate token from the vocabulary $\mathcal{V}$, and $\mathcal{T}{<k}$ represents the complete decoding context preceding the $k$-th entity token. If a predicted entity belongs to the Correct Matches set ($\mathbf{e} \in \mathcal{P}^{\text{CM}}$) and its uncertainty score exceeds a predefined threshold $\theta$, it is flagged as an \textit{uncertain entity}.

To suppress these latent risks, we construct an additional preference quadruplet $(x, \hat{x}^{\text{unc}}, \overline{y}^{\text{GT}}, \overline{y}^{\text{unc}})$ to execute M$^2$DPO.
Here, the report $\overline{y}^{\text{unc}}$ is derived via counterfactual modifications founded on $\overline{y}^{\text{GT}}$.
Specifically, we replace each uncertain entity with the second most probable candidate entity from the predictive distribution.
To ensure clinical plausibility, we introduce a knowledge graph verification based on RadGraph~\cite{jain2021radgraph}: if the sentence containing the substitute entity includes other medical entities, the substitute must share valid connection edges with them within the RadGraph schema.
$x^{\text{unc}}$ is then acquired using the retrieval protocol detailed in Sec.~\ref{sec:method_image_retrival}.
Finally, the optimization objective for mitigating these latent risks is formulated as $\mathcal{L}_{\text{M$^2$DPO}}(x, \hat{x}^{\text{unc}}, \overline{y}^{\text{GT}}, \overline{y}^{\text{unc}})$.

\subsection{Training Pipeline}
\label{sec:method_training_pipeline}
The overall training pipeline of DPO-Clin optimizes a joint multi-modal objective that simultaneously mitigates both explicit errors and latent risks.
To ensure stable convergence and strict clinical plausibility, we introduce dynamic binary indicators $\mathbbm{1}_{\text{EE}}$ and $\mathbbm{1}_{\text{LR}}$ to selectively mask invalid training instances.
The total training objective is formulated as:
\begin{equation}
\small
\mathcal{L}_{\text{total}} = 
\underbrace{\mathbbm{1}_{\text{EE}}\mathcal{L}_{\text{M$^2$DPO}}(x,\hat{x}^{\text{pre}},\overline{y}^{\text{GT}},y^{\text{pre}})}_{\text{mitigating explicit errors (Sec.~\ref{sec:method_image_retrival})}} + 
\underbrace{\mathbbm{1}_{\text{LR}}\mathcal{L}_{\text{M$^2$DPO}}(x,\hat{x}^{\text{unc}},\overline{y}^{\text{GT}},\overline{y}^{\text{unc}})}_{\text{mitigating latent risks (Sec.~\ref{sec:method_counterfactual_data})}}
\label{eq:total_objective}
\end{equation}
Here, $\mathbbm{1}_{\text{EE}} = 0$ if either the ECD module identifies no factual errors in $y^{\text{pre}}$ or the candidate pool obtained in the first-stage retrieval is empty. Conversely, $\mathbbm{1}_{\text{LR}} = 0$ if the counterfactual modifications are rejected by the knowledge graph verification. Otherwise, these indicators are set to $1$. This dynamic masking mechanism ensures the model exclusively learns from rigorously validated, high-quality multi-modal preference quadruplets.
 \section{Experiments}

\subsection{Dataset and Evaluation Metrics}
\noindent\textbf{Dataset.}
To comprehensively evaluate the efficacy of our DPO-Clin in MRG tasks, we conduct experiments on three MRG datasets: two widely-used chest X-ray (CXR) datasets, IU X-Ray~\cite{demner2015preparing} and MIMIC-CXR~\cite{johnson2019mimic}, and one in-house endoscopy dataset.
For the public CXR datasets, following the previous works~\cite{jin2024promptmrg,hou2025radar}, we adopt the official split of MIMIC-CXR, allocating $270,790$, $2,130$, and $3,858$ samples for training, validation, and testing, respectively.
To assess the model's generalization capability, the entire IU X-Ray dataset (totaling $3,955$ samples) is exclusively reserved for testing.
Regarding the in-house endoscopy dataset, we divide it into $3,925$ samples for training, $982$ for validation, and $982$ for testing.

\noindent\textbf{Evaluation Metrics.}
Following the previous works~\cite{hou2025radar,liu2025rrg} , we evaluate model performance using four natural language generation (NLG) metrics (BLEU-$1$/$4$~\cite{papineni2002bleu}, METEOR~\cite{banerjee2005meteor}, ROUGE-L~\cite{lin2004rouge}), and two clinical efficacy (CE) metrics (macro and micro $\mathrm{F_1}$ scores, $^{14}$Ma-$\mathrm{F_1}$ and $^{14}$Mi-$\mathrm{F_1}$), which are calculated by converting reports into $14$ abnormality classification labels using CheXbert~\cite{smit2020chexbert}.
Moreover, we employ three radiology report generation (RRG)-exclusive metrics (RadGraph~\cite{jain2021radgraph}, RadCliQ~\cite{yu2023evaluating}, RaTEScore~\cite{zhao2024ratescore}) for a more comprehensive evaluation.
Notably, as CheXbert is not suitable for endoscopy reports, the CE evaluation is restricted for binary diagnoses (normal/abnormal) on the endoscopy dataset, thus we employ the CE metric $^{2}\mathrm{F_1}$.

\subsection{Implementation Details}
During the SFT phase, we train the baseline models (R2GenGPT~\cite{wang2023r2gengpt} and RADAR~\cite{hou2025radar}) following their official supervision configurations.
During the DPO-Clin post-training phase, we employ Low-Rank Adaptation (LoRA)~\cite{hu2022lora} to efficiently fine-tune all linear layer parameters.
The model is optimized using the AdamW optimizer~\cite{loshchilov2017decoupled} with a weight decay of $0$, accompanied by a cosine scheduler.
Specifically, the training spans $5$ epochs with an initial learning rate of $1 \times 10^{-5}$ for the MIMIC-CXR dataset, and $20$ epochs with an initial learning rate of $2 \times 10^{-5}$ for the in-house endoscopy dataset.
Regarding DPO-specific configurations, the temperature hyperparameter $\beta$ is kept constant at $0.1$.
The empirical hyperparameters, specifically the matching threshold $\tau$ and the uncertainty threshold $\theta$, are set to $0.4$ and $0.5$ for the CXR datasets, and $0.5$ and $0.5$ for the endoscopy dataset, respectively.
Notably, when applying DPO-Clin to the in-house endoscopy dataset, the RaTE-NER model utilized in the ECD module is fine-tuned using data annotated by endoscopists, ensuring domain-accurate medical entity extraction.
Moreover, we employ GastroNet-5M~\cite{jong2025gastronet} to extract spatial vision embeddings during the image-based visual retrieval stage.
All experiments are conducted on $4$ NVIDIA RTX 4090 GPUs.

\subsection{Main Results}
\noindent\textbf{Baselines.}
We compare our DPO-Clin with three DPO-based VLM post-training methods, including a general-domain method, SIMA~\cite{wang2025enhancing}, and two medically-tailored methods, MMedPO~\cite{zhummedpo} and RRG-DPO~\cite{liu2025rrg}.
To evaluate universality across architectures, we employ two distinct baselines: R2GenGPT~\cite{wang2023r2gengpt}, representing a canonical VLM architecture, and RADAR~\cite{hou2025radar}, a SOTA MRG method that integrates an auxiliary expert model to augment clinical observations.
For a fair comparison, we implement a unified training pipeline, applying these preference optimization algorithms as a post-training stage following the supervised fine-tuned baseline.

\noindent\textbf{Comparison on two CXR Datasets.}
\begin{table}[!t]
\centering
\caption{Comparison of DPO-Clin with other DPO-based MRG methods on two CXR datasets, IU X-Ray and MIMIC-CXR. \textbf{Bold} and \underline{underlined} text indicates the best performance and second-best performance, respectively.}
\resizebox{\textwidth}{!}{
\begin{tabular}{c|l|cccc|cc|ccc}
\toprule
\rowcolor[HTML]{ECECEC} &  & \multicolumn{4}{c|}{\textbf{NLG Metrics}} & \multicolumn{2}{c|}{\textbf{CE Metrics}} & \multicolumn{3}{c}{\textbf{RRG-exclusive Metrics}}\\ 
\rowcolor[HTML]{ECECEC}\multirow{-2}{*}{\textbf{Dataset}} & \multirow{-2}{*}{\textbf{Methods}} & \textbf{B-$1$ \textuparrow} & \textbf{B-$4$ \textuparrow} & \textbf{MTR \textuparrow} & \textbf{R-L \textuparrow} & {$\mathbf{^{14}}$\textbf{Ma-}$\mathbf{F_1}$ \textuparrow} & {$\mathbf{^{14}}$\textbf{Mi-}$\mathbf{F_1}$ \textuparrow} & \textbf{RadGraph \textuparrow} &\textbf{RadCliQ \textdownarrow}& \textbf{RaTEScore \textuparrow} \\ 
\midrule \midrule
\multirow{10}{*}{\makecell{\textbf{MIMIC-CXR} \\ (\textit{Training} \& \\ \textit{Evaluation})}}
& R2GenGPT & 0.411 & 0.134 & 0.160 & 0.297 & 0.389 & 0.504 & 0.260 & 2.74 & 0.453 \\
& +~SIMA & 0.416 & \underline{0.140} & 0.168 & 0.306 & 0.396 & 0.515 & 0.265 & 2.74 & 0.450 \\
& +~MMedPO & \textbf{0.426} & \textbf{0.144} & \textbf{0.174} & \underline{0.315} & \underline{0.420} & \underline{0.530} & \underline{0.281} & \underline{2.71} & \underline{0.476} \\
& +~RRG-DPO & 0.417 & 0.138 & 0.165 & 0.302 & 0.415 & 0.522 & 0.275 & \underline{2.71} & 0.470 \\
& \cellcolor[HTML]{E9F3FE}\textbf{+~Ours} & \cellcolor[HTML]{E9F3FE}\underline{0.423} & \cellcolor[HTML]{E9F3FE}\underline{0.140} & \cellcolor[HTML]{E9F3FE}\textbf{0.174} & \cellcolor[HTML]{E9F3FE}\textbf{0.321} & \cellcolor[HTML]{E9F3FE}\textbf{0.437} & \cellcolor[HTML]{E9F3FE}\textbf{0.545} & \cellcolor[HTML]{E9F3FE}\textbf{0.297} & \cellcolor[HTML]{E9F3FE}\textbf{2.67} & \cellcolor[HTML]{E9F3FE}\textbf{0.494}\\
\cline{2-11}
& RADAR & 0.509 & 0.262 & 0.450 & 0.397 & 0.460 & 0.627 & 0.346 & 2.61 & 0.531 \\
& +~SIMA & 0.512 & 0.264 & 0.459 & 0.405 & 0.464 & 0.620 & 0.341 & 2.57 & 0.537 \\
& +~MMedPO & \underline{0.516} & \underline{0.268} & \underline{0.466} & \underline{0.410} & \underline{0.478} & \underline{0.639} & \underline{0.356} & \underline{2.52} & 0.546 \\
& +~RRG-DPO & 0.511 & 0.260 & 0.459 & 0.408 & 0.473 & \underline{0.639} & 0.356 & 2.54 & \underline{0.550} \\
& \cellcolor[HTML]{E9F3FE}\textbf{+~Ours} & \cellcolor[HTML]{E9F3FE}\textbf{0.522} & \cellcolor[HTML]{E9F3FE}\textbf{0.271} & \cellcolor[HTML]{E9F3FE}\textbf{0.471} & \cellcolor[HTML]{E9F3FE}\textbf{0.419} & \cellcolor[HTML]{E9F3FE}\textbf{0.490} & \cellcolor[HTML]{E9F3FE}\textbf{0.654} & \cellcolor[HTML]{E9F3FE}\textbf{0.372} & \cellcolor[HTML]{E9F3FE}\textbf{2.48} & \cellcolor[HTML]{E9F3FE}\textbf{0.567}\\
\midrule
\multirow{10}{*}{\makecell{\textbf{IU X-Ray} \\ (\textit{Evaluation})}}
& R2GenGPT & 0.356 & 0.103 & 0.138 & 0.270 & 0.303 & 0.446 & 0.204 & 2.86 & 0.405 \\
& +~SIMA & 0.359 & 0.106 & 0.143 &0.277 & 0.312 & 0.458 & 0.210 & 2.85 & 0.413 \\
& +~MMedPO & \textbf{0.370} & \textbf{0.114} & \underline{0.153} & \underline{0.286} & \underline{0.332} & \underline{0.475} & \underline{0.231} & \underline{2.80} & \underline{0.428} \\
& +~RRG-DPO & \underline{0.367} & 0.110 & 0.146 & 0.277 & 0.326 & 0.464 & 0.220 & 2.82 & 0.422 \\
& \cellcolor[HTML]{E9F3FE}\textbf{+~Ours} & \cellcolor[HTML]{E9F3FE}{0.365} & \cellcolor[HTML]{E9F3FE}\underline{0.112} & \cellcolor[HTML]{E9F3FE}\textbf{0.159} & \cellcolor[HTML]{E9F3FE}\textbf{0.292} & \cellcolor[HTML]{E9F3FE}\textbf{0.350} & \cellcolor[HTML]{E9F3FE}\textbf{0.492} & \cellcolor[HTML]{E9F3FE}\textbf{0.252} & \cellcolor[HTML]{E9F3FE}\textbf{2.74} & \cellcolor[HTML]{E9F3FE}\textbf{0.452}\\
\cline{2-11}
& RADAR & 0.364 & 0.116 & 0.142 & 0.276 & 0.325 & 0.546 & 0.237 & 2.78 & 0.428 \\
& +~SIMA & 0.370 & 0.118 & 0.148 & 0.281 & 0.331 & 0.555 & 0.244 & 2.76 & 0.435 \\
& +~MMedPO & \underline{0.374} & \textbf{0.126} & \underline{0.152} & \underline{0.289} & \underline{0.345} & \underline{0.568} & \underline{0.259} & \underline{2.70} & \underline{0.452} \\
& +~RRG-DPO & 0.372 & 0.124 & 0.148 & 0.285 & 0.338 & 0.560 & 0.251 & 2.72 & 0.444 \\
& \cellcolor[HTML]{E9F3FE}\textbf{+~Ours} & \cellcolor[HTML]{E9F3FE}\textbf{0.377} & \cellcolor[HTML]{E9F3FE}\textbf{0.126} & \cellcolor[HTML]{E9F3FE}\textbf{0.160} & \cellcolor[HTML]{E9F3FE}\textbf{0.298} & \cellcolor[HTML]{E9F3FE}\textbf{0.362} & \cellcolor[HTML]{E9F3FE}\textbf{0.580} & \cellcolor[HTML]{E9F3FE}\textbf{0.275} & \cellcolor[HTML]{E9F3FE}\textbf{2.66} & \cellcolor[HTML]{E9F3FE}\textbf{0.461}\\
\bottomrule
\end{tabular}
}
\label{tab:com_SOTA_CXR}
\end{table}
In Table~\ref{tab:com_SOTA_CXR}, we validate the in-domain performance on MIMIC-CXR and cross-domain generalization on IU X-Ray, respectively.
It can be seen that all of these preference optimization algorithms consistently enhance the baseline performance on both datasets, demonstrating the effectiveness of preference optimization following supervised fine-tuning.
Crucially, our DPO-Clin outperforms competing methods in almost all metrics, showing the most significant improvements in report quality, particularly in CE and RRG-exclusive metrics that emphasize clinical relevance.
These results confirm the superiority of our DPO-Clin in enforcing clinical factuality and generating medically accurate reports.

\noindent\textbf{Comparison on the Endoscopy Dataset.}
\begin{table*}[t!]
\centering
\caption{Comparison on the in-house endoscopy report dataset.}
\setlength{\tabcolsep}{14.3 pt}
\scriptsize
\begin{tabular}{l|cccc|c}
\toprule
\rowcolor[HTML]{ECECEC} & \multicolumn{4}{c|}{\textbf{NLG Metrics}} & {\textbf{CE Metric}} \\
\rowcolor[HTML]{ECECEC} \multirow{-2}{*}{\textbf{Methods}} & \textbf{B-$1$} & \textbf{B-$4$} & \textbf{MTR} & \textbf{R-L} & $\mathbf{^2F_1}$ \\ 
\midrule \midrule
R2GenGPT & 0.528 & 0.401 & 0.545 & 0.523 & 0.798 \\
+~SIMA & 0.546 & 0.418 & 0.574 & 0.541 & 0.825 \\
+~MMedPO & \underline{0.558} & \underline{0.424} & \underline{0.585} & \underline{0.557} & \underline{0.834} \\
+~RRG-DPO & 0.540 & 0.417 & 0.566 & 0.536 & 0.813 \\
\cellcolor[HTML]{E9F3FE}\textbf{+~Ours} & \cellcolor[HTML]{E9F3FE}\textbf{0.576} & \cellcolor[HTML]{E9F3FE}\textbf{0.435} & \cellcolor[HTML]{E9F3FE}\textbf{0.607} & \cellcolor[HTML]{E9F3FE}\textbf{0.574} & \cellcolor[HTML]{E9F3FE}\textbf{0.852} \\
\bottomrule
\end{tabular}
\label{tab:com_SOTA_colon}
\end{table*}
To further validate the universality of different methods beyond the radiology scenario, we conduct experiments by training models from scratch on the in-house endoscopy report dataset.
R2GenGPT is selected as the baseline model as it features a streamlined, modality-agnostic architecture that enables direct transfer to endoscopy scenario.
As illustrated in Table~\ref{tab:com_SOTA_colon}, the comparison results on the endoscopy dataset align with those observed in the CXR benchmark.
Compared to other preference optimization methods, our DPO-Clin consistently maintains a significant performance advantage.
Specifically, DPO-Clin achieves the highest CE score, with its $^2\text{F}_1$ value improving by $5.4\%$ ($0.852$ \textit{vs.} $0.798$) compared to the baseline model and surpassing the second-best method MMedPO by $1.8\%$ ($0.852$ \textit{vs.} $0.834$).
This demonstrates DPO's robustness and effectiveness across diverse medical imaging modalities.

\subsection{In-Depth Analysis}
\subsubsection{Effectiveness of Mitigating Explicit Errors and Latent Risks.}
\begin{table}[!t]
\centering
\caption{Ablation on the effectiveness of Mitigating Explicit Errors and Latent Risks, denoted as `MEE' and `MLR'. `ECD', `Verification', and `Masking' denote Entity-level Clinical Diagnostic module, knowledge graph verification, and dynamic masking mechanism, respectively. `$\overline{y}^{\text{GT}} \rightarrow y^{\text{GT}}$' indicates using the GT report $y^{\text{GT}}$ instead of the generated linguistically-aligned report $\overline{y}^{\text{GT}}$ in the total objective $\mathcal{L}_{\text{total}}$.}
\resizebox{\textwidth}{!}{
\begin{tabular}{l|cccc|cc|ccc}
\toprule
\rowcolor[HTML]{ECECEC} & \multicolumn{4}{c|}{\textbf{NLG Metrics}} & \multicolumn{2}{c|}{\textbf{CE Metrics}} & \multicolumn{3}{c}{\textbf{RRG-exclusive Metrics}}\\ 
\rowcolor[HTML]{ECECEC} \multirow{-2}{*}{\textbf{Methods}} & \textbf{B-}$\mathbf{1}$ & \textbf{B-}$\mathbf{4}$ & \textbf{MTR} & \textbf{R-L} & $\mathbf{^{14}}$\textbf{Ma-}$\mathbf{F_1}$ & $\mathbf{^{14}}$\textbf{Mi-}$\mathbf{F_1}$ & \textbf{RadGraph} &\textbf{RadCliQ}& \textbf{RaTEScore} \\ 
\midrule \midrule
RADAR (Baseline) & 0.509 & 0.262 & 0.450 & 0.397 & 0.460 & 0.627 & 0.346 & 2.61 & 0.531 \\
\midrule
+~MEE (\textit{w/o} ECD) & 0.498 & 0.256 & 0.445 & 0.390 & 0.452 & 0.615 & 0.337 & 2.64 & 0.518 \\
+~MEE & 0.518 & 0.268 & 0.466 & 0.414 & 0.482 & 0.645 & 0.366 & 2.51 & 0.557 \\
\midrule
+~MLR (\textit{w/o} Verification) & 0.513 & 0.265 & 0.454 & 0.406 & 0.472 & 0.635 & 0.354 & 2.55 & 0.546 \\
+~MLR & 0.513 & 0.266 & 0.459 & 0.409 & 0.475 & 0.640 & 0.360 & 2.55 & 0.550 \\
\midrule
+~MEE\&MLR (\textit{w/o} Masking) & 0.512 & 0.262 & 0.454 & 0.403 & 0.466 & 0.638 & 0.353 & 2.58 & 0.541 \\
+~MEE\&MLR ($\overline{y}^{\text{GT}} \rightarrow y^{\text{GT}}$) & \textbf{0.526} & \textbf{0.273} & 0.466 & 0.417 & 0.482 & 0.642 & 0.360 & 2.52 & 0.552 \\
\cellcolor[HTML]{E9F3FE}\textbf{+~MEE\&MLR (DPO-Clin)} & \cellcolor[HTML]{E9F3FE}{0.522} & \cellcolor[HTML]{E9F3FE}{0.271} & \cellcolor[HTML]{E9F3FE}\textbf{0.471} & \cellcolor[HTML]{E9F3FE}\textbf{0.419} & \cellcolor[HTML]{E9F3FE}\textbf{0.490} & \cellcolor[HTML]{E9F3FE}\textbf{0.654} & \cellcolor[HTML]{E9F3FE}\textbf{0.372} & \cellcolor[HTML]{E9F3FE}\textbf{2.48} & \cellcolor[HTML]{E9F3FE}\textbf{0.567}\\
\bottomrule
\end{tabular}
}
\label{tab:ablation_MEE_MLR}
\end{table}

In Table~\ref{tab:ablation_MEE_MLR}, we progressively augment the RADAR baseline with our proposed components on the MIMIC-CXR dataset. The results demonstrate that mitigating explicit textual errors and latent clinical risks independently can further improve the quality of medical reports generated by the baseline model. Moreover, the combination of both components achieves the best performance, indicating that their contributions to the baseline optimization are complementary.

Importantly, the two proposed sub-components, Entity-level Clinical Diagnostic (ECD) module and Knowledge Graph Verification mechanism, both play a pivotal role in ensuring clinical reliability.
As shown in Table~\ref{tab:ablation_MEE_MLR}, disabling the verification mechanism yields limited improvements, while removing the ECD module results in performance degradation compared to the baseline. This discrepancy arises primarily because the ECD module guarantees that the LLM-generated reports $\overline{y}_\text{GT}$ adhere to clinical factuality, whereas the verification ensures the clinical plausibility of the counterfactually modified reports $\overline{y}^{\text{unc}}$.

Furthermore, the ablation underscores the necessity of the linguistically-aligned preference construction and the dynamic masking mechanism within the joint training framework. As shown in the last three rows of Table~\ref{tab:ablation_MEE_MLR}, substituting the LLM-generated linguistically-aligned reports $\overline{y}^{\text{GT}}$ with the raw GT reports $y^{\text{GT}}$ marginally improves $n$-gram metrics (\textit{i.e.}, BLEU-1 and BLEU-4) but significantly degrades CE and RRG-exclusive metrics.
This indicates that while the model learns to mimic clinically irrelevant descriptions, it does so at the severe cost of clinical accuracy.
Such a trade-off shows the necessity of constructing linguistically-aligned preference data to ensure the optimization focuses on capturing clinically relevant discrepancies.
Similarly, disabling the dynamic masking mechanism compromises the overall framework performance. It ensures that the optimization targets failed predictions, as well as the reliability of the retrieved images $\hat{x}^{\text{GT}}$ and $\hat{x}^{\text{unc}}$ and the counterfactually modified reports $\overline{y}^{unc}$, thereby maintaining the stability and efficacy of the entire pipeline.

\subsubsection{Comparison of Different DPO Variants.}
\begin{table}[!t]
\centering
\caption{Performance comparison of Different DPO Variants.}
\setlength{\tabcolsep}{4.5 pt}
\scriptsize
\begin{tabular}{l|ccccc|cc}
\toprule
\rowcolor[HTML]{ECECEC} & \multicolumn{5}{c|}{\textbf{MIMIC-CXR}} & \multicolumn{2}{c}{\textbf{Endoscopy}}\\
\rowcolor[HTML]{ECECEC} & {\textbf{NLG}} & {\textbf{CE}} & \multicolumn{3}{c|}{\textbf{RRG-exclusive}} &{\textbf{NLG}} & {\textbf{CE}} \\
\rowcolor[HTML]{ECECEC} \multirow{-3}{*}{\textbf{Methods}} & \textbf{R-L} & $\mathbf{^{14}}$\textbf{Ma-}$\mathbf{F_1}$ & \textbf{Graph} &\textbf{CliQ} &\textbf{RaTEScore} &\textbf{R-L} & $\mathbf{^{2}F_1}$ \\
\midrule \midrule
R2GenGPT (Baseline) & 0.297 & 0.389 & 0.260 & 2.74 & 0.453 & 0.523 & 0.798 \\
\midrule
+~standard DPO & 0.311  & 0.420 & 0.278 & 2.71 & 0.471 & 0.549 & 0.832 \\
+~SimPO & 0.313 & 0.425 & 0.287 & 2.70 & 0.480 & 0.560 & 0.838  \\
\midrule
\cellcolor[HTML]{E9F3FE}\textbf{+~M$^2$DPO (Ours)} & \cellcolor[HTML]{E9F3FE}\textbf{0.321} & \cellcolor[HTML]{E9F3FE}\textbf{0.437} & \cellcolor[HTML]{E9F3FE}\textbf{0.297} & \cellcolor[HTML]{E9F3FE}\textbf{2.67} & \cellcolor[HTML]{E9F3FE}\textbf{0.494} & \cellcolor[HTML]{E9F3FE}\textbf{0.574} & \cellcolor[HTML]{E9F3FE}\textbf{0.852} \\
\bottomrule
\end{tabular}
\label{tab:ablation_DPO_Variants}
\end{table}
To validate the effectiveness of incorporating visual constraints during preference optimization, we compare our M$^2$DPO with text-centric competitors, the standard DPO and SimPO, in Table \ref{tab:ablation_DPO_Variants}.
We conduct evaluation on MIMIC-CXR and the in-house endoscopy dataset, setting R2GenGPT as the baseline model.
Taking the standard DPO implementation as an example, the overall objective in Eq.~(\ref{eq:total_objective}) is reformulated as $\mathcal{L}_{\text{total}}=\mathbbm{1}_{\text{EE}}\mathcal{L}_{\text{DPO}}(x,\overline{y}^{\text{GT}},y^{\text{pre}}) + \mathbbm{1}_{\text{LR}}\mathcal{L}_{\text{DPO}}(x,\overline{y}^{\text{GT}},\overline{y}^{\text{unc}})$.
As illustrated in Table~\ref{tab:ablation_DPO_Variants}, our M$^2$DPO consistently outperforms both DPO and SimPO across all evaluated metrics on both datasets. This quantitatively confirms that incorporating a multi-modal preference alignment mechanism enhances the clinical reliability of MRG models across diverse medical imaging modalities.

\begin{figure}[htbp]
    \centering
 
    \begin{minipage}{0.43\textwidth}
        \centering
        \includegraphics[width=\linewidth]{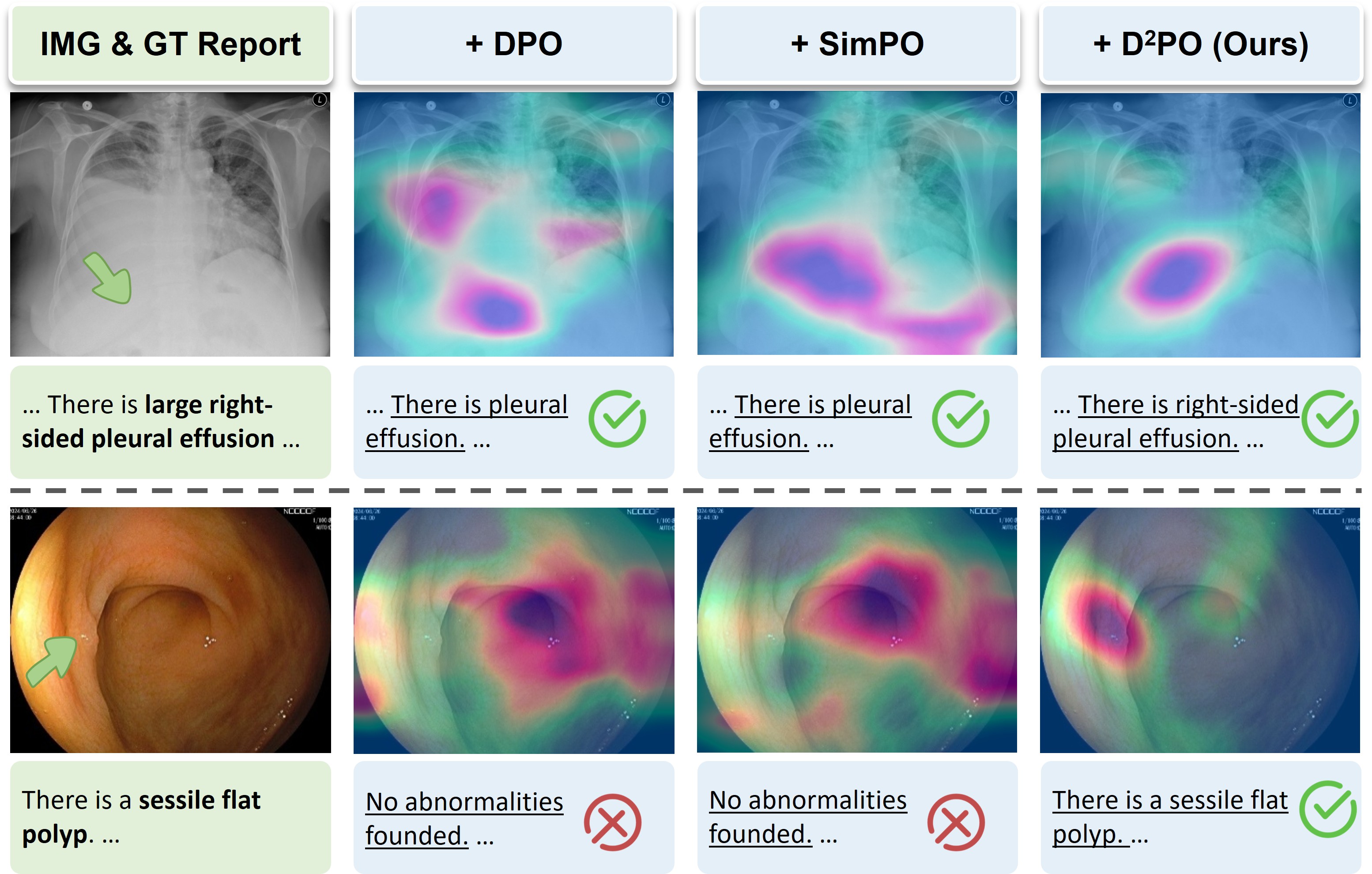}
        \caption{Visualization of cross-modal attention maps. The attention map represents the average attention weights of all tokens within the underlined sentence.}
        \label{fig:attn_visualization}
    \end{minipage}
    \hfill
    \begin{minipage}{0.54\textwidth}
        \centering
        \includegraphics[width=\linewidth]{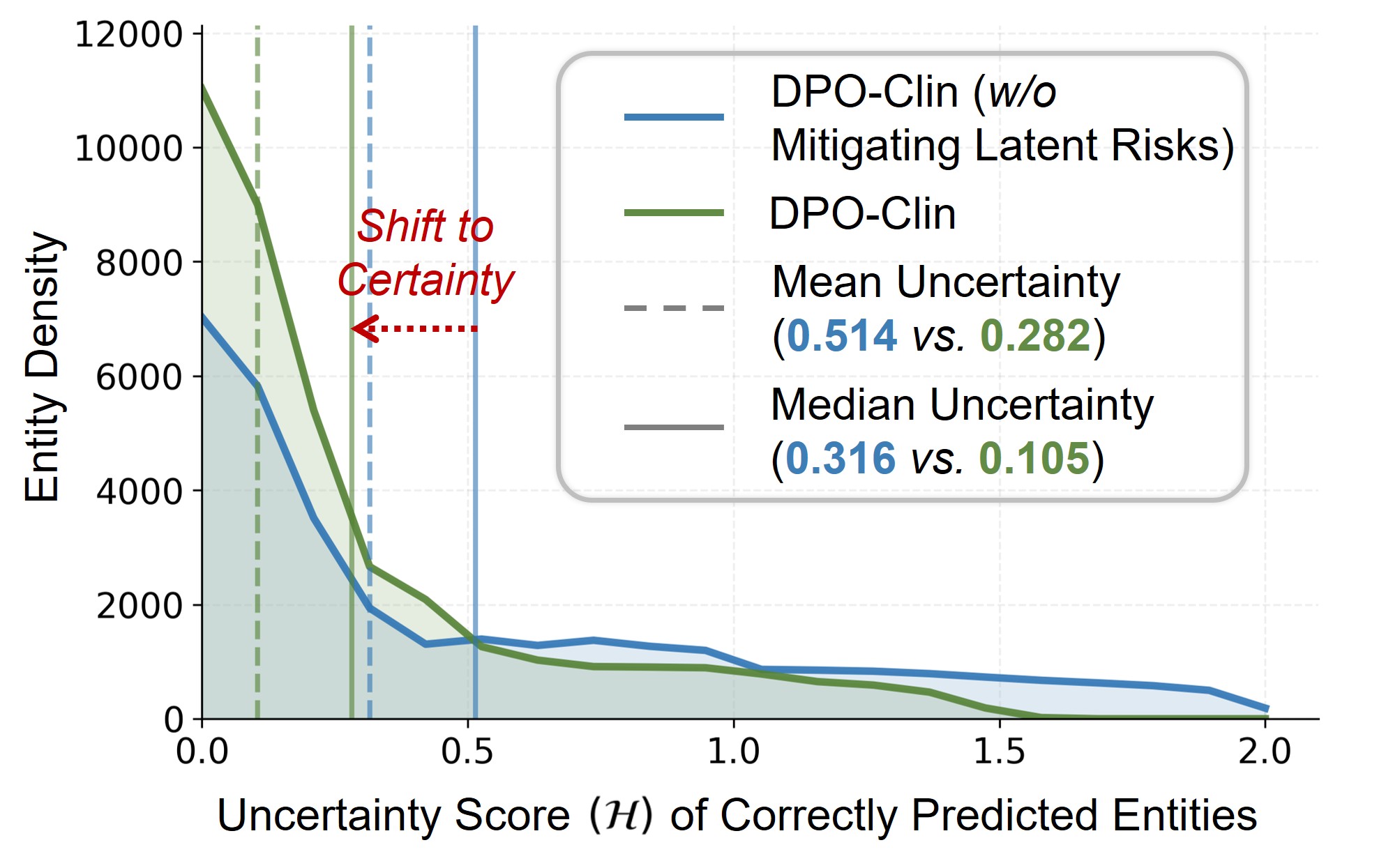} 
        \caption{Density distribution of uncertainty scores for correctly predicted entities with/without mitigating latent risks.}
        \label{fig:exp_mitigate_LH}
    \end{minipage}

\end{figure}

To further analyze the underlying mechanism, we visualize the cross-modal attention weights between the generated text tokens and the input visual tokens in the LLM decoder.
As shown in Fig.~\ref{fig:attn_visualization}, models optimized with the standard DPO and SimPO both exhibit scattered attention distribution that are inconsistent with the actual pathological regions (indicated by green arrows).
This visual misalignment results in ambiguous descriptions (\textit{e.g.}, the generic `pleural effusion') or severe factual errors (\textit{e.g.}, missing of the polyp).
In contrast, M$^2$DPO yields attention maps that are more consistent with the actual lesion regions, which empowers the model to tackle challenging scenarios, such as detecting the small sessile flat polyp.
These results verify that M$^2$DPO enforces fine-grained vision-language alignment of the MRG baselines.

\subsubsection{Effect of Latent Risks Mitigation on Entity Uncertainty.}
As illustrated in Fig.~\ref{fig:exp_mitigate_LH}, we plot the frequency distribution curves of uncertainty scores for all correctly predicted entities on MIMIC-CXR.
Using RADAR as the baseline model, the entity extraction and correctness verification are performed via the ECD module, with uncertainty scores calculated according to Eq.~(\ref{eq:uncertainty_score}).
It is seen that after mitigating latent risks, the uncertainty distribution of correct entities exhibits a shift toward lower values (mean uncertainty score: $0.514 \rightarrow 0.282$, median uncertainty score: $0.316 \rightarrow 0.105$).
This `shift to certainty' phenomenon demonstrates the model's increasing confidence in its accurate predictions. Furthermore, the expansion of the area under the curve intuitively signifies an increase in the total count of correctly predicted entities, thus validating the enhanced clinical accuracy of the generated reports.

\section{Conclusion}
In this paper, we propose DPO-Clin, a DPO-based post-training framework tailored for MRG. Existing DPO-based MRG methods suffer from fundamental limitations: their preference curation inadvertently entangles critical clinical findings with clinically irrelevant linguistic characteristics, and they lack explicit vision-language alignment. To address these issues, we introduce the ECD module to assist in generating linguistically-aligned preference data, forcing the optimization process to focus exclusively on clinically relevant discrepancies. Furthermore, we propose M$^2$DPO to achieve fine-grained vision-language grounding. By explicitly emphasizing the necessity of mitigating latent risks, DPO-Clin further elevates the diagnostic reliability of the generated reports. Extensive evaluations across diverse medical imaging modalities (chest X-rays and endoscopy) consistently demonstrate that DPO-Clin achieves SOTA performance. Furthermore, the plug-and-play generalizability of DPO-Clin has also been verified across two distinct baseline architectures (R2GenGPT and RADAR), paving the way for groundbreaking advancements in the future development of trustworthy medical report generation. This is conducive to the further development of trustworthy medical report generation.

\section*{Acknowledgements}
This work was supported in part by National Natural Science Foundation of China (Grant No.62576145), research grants from Wuhan United Imaging Healthcare Surgical Technology Co., Ltd.

%
%
\bibliographystyle{splncs04}
\bibliography{reference}


\definecolor{placeholderbg}{RGB}{255, 248, 210}
\definecolor{placeholderfg}{RGB}{180, 120, 0}
\newcommand{\placeholder}[1]{\colorbox{placeholderbg}{\textcolor{placeholderfg}{#1}}}

\newtcolorbox{promptbox}[1][]{
    enhanced,
    breakable,
    colback=gray!4,
    colframe=black!65,
    boxrule=0.6pt,
    arc=4pt,
    left=10pt, right=10pt, top=6pt, bottom=14pt,
    before upper={\footnotesize\setlength{\parskip}{4pt}},
    fonttitle=\bfseries\small,
    title=#1,
    extras middle and last={top=18pt},
    overlay first and middle={
        \node[anchor=south east, inner sep=4pt,
              font=\footnotesize\itshape, text=black!45]
            at (frame.south east) {Continued on next page\ldots};
    },
    overlay middle and last={
        \node[anchor=north west, inner sep=6pt,
              font=\footnotesize\itshape, text=black!45]
            at (frame.north west) {(Continued from previous page)};
    }
}

\newcommand{\promptsec}[1]{%
  \par\vspace{4pt}%
  {\setlength{\fboxsep}{5pt}%
   \noindent\colorbox{black!10}{\makebox[\linewidth][l]{\bfseries\scriptsize #1}}}%
  \par\vspace{3pt}%
}


\appendix

\hypersetup{linkcolor=citecolor}
\hypersetup{urlcolor=blue, citecolor=citecolor, linkcolor=red}
\clearpage

\setcounter{page}{1}

\vspace*{1em} 
\noindent{\huge \bfseries Appendix}
\par\vspace{1em} 

\section{Hyperparameter Configuration}
\vspace{-0.9cm}
\begin{figure}
    \centering
    \includegraphics[width=\linewidth]{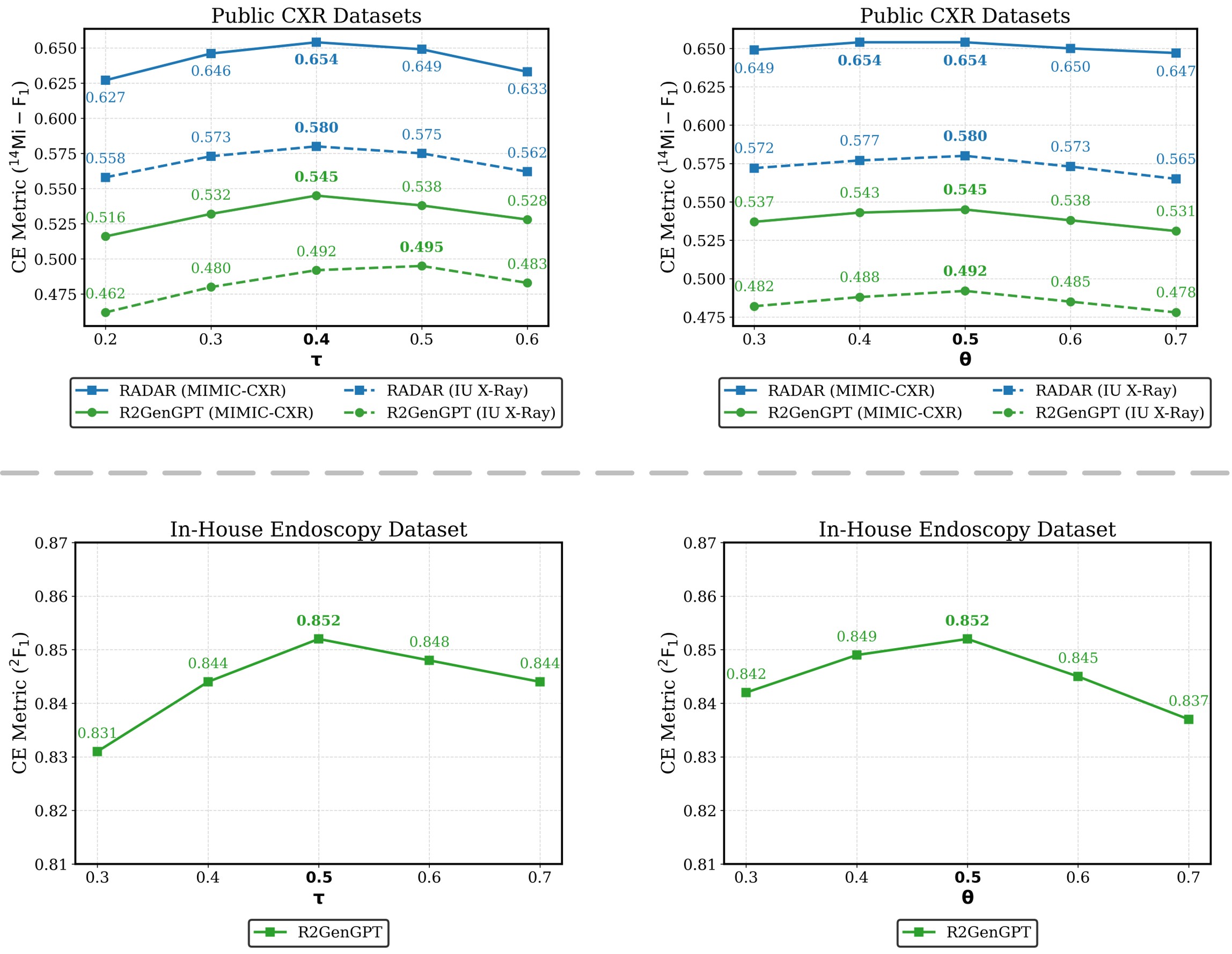}
    \vspace{-0.3cm}
    \caption{Ablation on two hyperparameters: a matching threshold $\tau$ and an uncertainty threshold $\theta$}
    \vspace{-0.5cm}
    \label{fig:hyperparameter}
\end{figure}

We conduct an ablation study on two Chest X-ray (CXR) datasets and an in-house endoscopy dataset to examine the effect of the values of two hyperparameters: a matching threshold $\tau$ (defined in Sec.~4.2 of the manuscript) and an uncertainty threshold $\theta$ (defined in Sec.~4.4 of the manuscript).
Specifically, $\tau$ governs the strictness of semantic entity matching within the Entity-level Clinical Diagnostic (ECD) module, while $\theta$ establishes the boundary for identifying highly uncertain entities.

As illustrated in Fig.~\ref{fig:hyperparameter}, we evaluate the clinical efficacy (CE) metrics ($\mathbf{^{14}}$\textbf{Mi-}$\mathbf{F_1}$ and $^{2}\mathbf{F_1}$) in search spaces for both hyperparameters. The performance curves consistently exhibit an inverted U-shape across different baseline models (RADAR and R2GenGPT) and imaging modalities. Lower threshold values tend to introduce noisy optimization signals by either forcing incorrect semantic matches or excessively flagging confident predictions as uncertain. Conversely, overly high thresholds severely restrict the preference optimization scope by rejecting valid entity pairs or overlooking genuine latent risks. Consequently, based on the optimal peak performances observed, we empirically set $\tau=0.4$ and $\theta=0.5$ for the public CXR datasets. For the in-house endoscopy dataset, both hyperparameters are configured to $0.5$.

\section{Prompt Template for LLM-based Linguistically-Aligned Preference Report Generation}

In this section, we provide the detailed prompt template for driving large language model (LLM, specifically GPT-4o) to generate linguistically-aligned preference reports (\textit{i.e.}, $\overline{y}^{\text{GT}}$, defined in Sec.~4.2 of the manuscript). The prompt effectively integrates the predicted report, the ground truth (GT) report, and the structured diagnostic information (\textit{i.e.}, $\mathcal{P}^{\text{CM}}$, $\mathcal{P}^{\text{EE}}$, $\mathcal{P}^{\text{ME}}$, $\mathcal{P}^{\text{FA}}$, as detailed in Sec.~4.2 of the manuscript) output from the ECD module.
The detailed prompt template is presented below:

\begin{promptbox}[Prompt Template for LLM-based Linguistically-Aligned Preference Report Generation]

\promptsec{System Prompt}

You are an expert radiologist and clinical text editor.
Your task is to rewrite the predicted report into a \emph{linguistically-aligned preferred report} that:
(1) is clinically faithful to the GT (Ground Truth) report, and
(2) preserves the wording style and sentence structure of the predicted report as much as possible.
Use the predicted report as the linguistic template, the GT report as the factual source of truth, and the ECD diagnostic information as explicit editing guidance.
Follow these rules:
\begin{itemize}[leftmargin=1.2em, itemsep=0pt, topsep=2pt, parsep=0pt]
    \item Preserve clinically correct content from the predicted report whenever possible.
    \item Remove unsupported findings in $\mathcal{P}_{EE}$.
    \item Insert missing GT findings in $\mathcal{P}_{ME}$.
    \item Correct assertion conflicts in $\mathcal{P}_{FA}$.
    \item Prefer minimal local edits over global rewriting.
    \item Do not introduce facts not supported by the GT report.
\end{itemize}

\promptsec{User Prompt}

Please generate a revised report whose clinical facts align with the \textit{GT Report}, while preserving the linguistic characteristics of the \textit{Predicted Report} as much as possible.

\vspace{2pt}
\textit{\textless GT\_Report\textgreater}\\
\placeholder{\texttt{[Insert GT Report Here]}}\\
\textit{\textless /GT\_Report\textgreater}

\vspace{3pt}
\textit{\textless Predicted\_Report\textgreater}\\
\placeholder{\texttt{[Insert Predicted Report Here]}}\\
\textit{\textless /Predicted\_Report\textgreater}

\vspace{3pt}
\noindent\begin{minipage}{\linewidth}
\textit{\textless ECD\_Diagnostic\_Info\textgreater}

\vspace{2pt}
\begin{tabular}{@{\hspace{2pt}}ll@{}}
    \textbf{Correct Matches:}     & \placeholder{\texttt{[Insert }$\mathcal{P}^{\text{CM}}$\texttt{ Here]}} \\[1.5pt]
    \textbf{Extraneous Entities:} & \placeholder{\texttt{[Insert }$\mathcal{P}^{\text{EE}}$\texttt{ Here]}} \\[1.5pt]
    \textbf{Missing Entities:}    & \placeholder{\texttt{[Insert }$\mathcal{P}^{\text{ME}}$\texttt{ Here]}} \\[1.5pt]
    \textbf{False Assertions:}    & \placeholder{\texttt{[Insert }$\mathcal{P}^{\text{FA}}$\texttt{ Here]}} \\
\end{tabular}
\vspace{2pt}

\textit{\textless /ECD\_Diagnostic\_Info\textgreater}
\end{minipage}

\promptsec{Output Constraint}

Return \textbf{only} the rewritten report and nothing else.

\end{promptbox}

\section{More Visualizations of Preference Data Construction}
To intuitively illustrate the operation mechanism of our DPO-Clin framework, we provide qualitative examples of the preference data construction process in Fig.~\ref{fig:case_study1} (Part I and Part II). Each case presents the input medical image, the initial predicted report $y^{\text{pre}}$, the GT report $y^{\text{GT}}$, the fine-grained diagnostic results obtained by the ECD module (categorized into four distinct entity sets $\{ \mathcal{P}^{\text{CM}}, \mathcal{P}^{\text{EE}}, \mathcal{P}^{\text{ME}}, \mathcal{P}^{\text{FA}}\}$), and the final linguistically-aligned report $\overline{y}^{\text{GT}}$ generated by the LLM. These examples demonstrate how the LLM seamlessly calibrates factual errors while preserving the original linguistic characteristics.
\begin{figure}[h]
    \centering
    \includegraphics[width=0.95\linewidth]{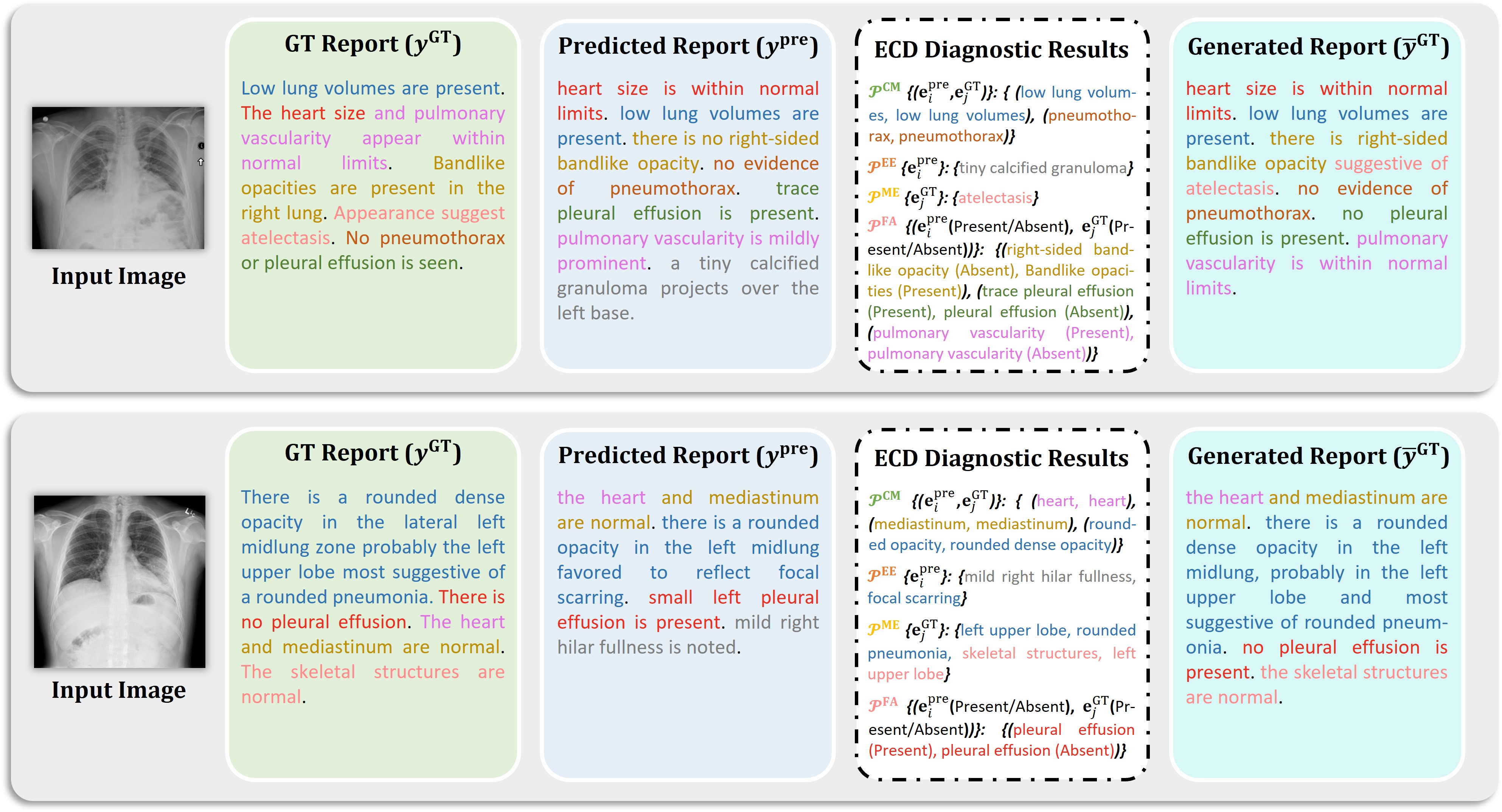}
    \caption{Visualization of linguistically-aligned preference reports generation (Part \textbf{I}). Distinct colors are used to indicate words corresponding to different clinical
findings.}
    \label{fig:case_study1}
\end{figure}

\begin{figure}[h]
    \centering
    \ContinuedFloat 
    \includegraphics[width=0.95\textwidth]{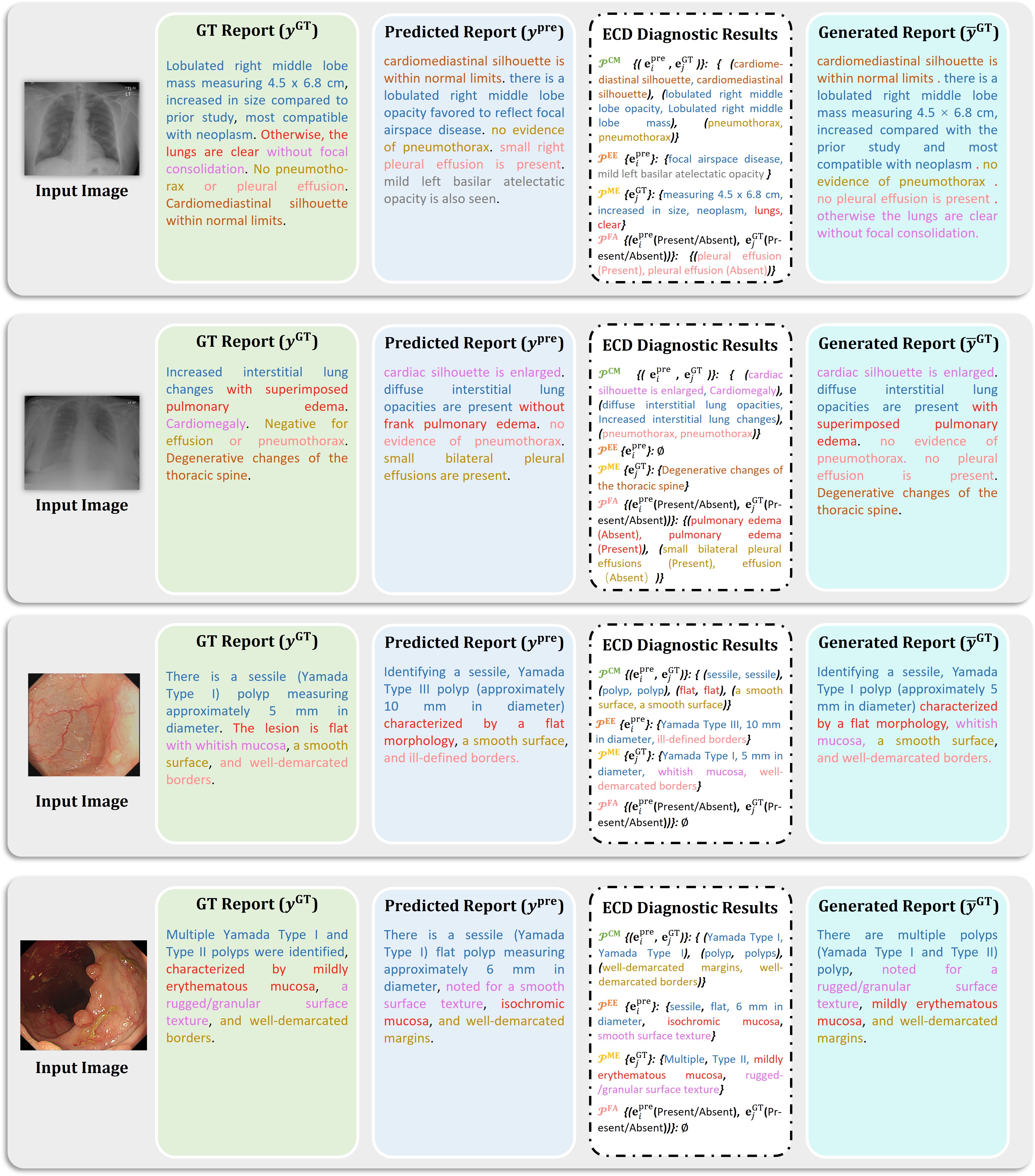}
    \caption{Visualization of linguistically-aligned preference reports generation (Part \textbf{II}). Distinct colors are used to indicate words corresponding to different clinical
findings.}
\label{fig:case_study2}
\end{figure}



\end{document}